\documentclass[10pt,twocolumn,letterpaper]{article}
\usepackage[pagenumbers]{cvpr}

\usepackage{arydshln}
\usepackage{booktabs,array}
\usepackage{colortbl}
\usepackage{tabularx}
\usepackage{multirow}
\usepackage{wrapfig}
\usepackage{graphicx}
\usepackage{amsmath,amssymb,amsfonts}
\usepackage{algorithmic}
\usepackage{graphicx}
\usepackage{textcomp}
\usepackage[table,x11names,dvipsnames,table]{xcolor}
\usepackage{color} 
\usepackage{soul}
\usepackage{xspace}
\usepackage{balance}
\usepackage{bm}
\usepackage[skip=8pt]{caption}
\newcommand{\bb}[1]{\textbf{#1}}
\usepackage{bbding}
\newcolumntype{C}[1]{>{\centering\arraybackslash}p{#1}}

\newcommand{\NAME}{{QuAD}}
\newcommand{\NAMEvirali}{{ReWIND}}
\newcommand{\NAMEinlab}{{AncesTree}}

\definecolor{cvprblue}{rgb}{0.21,0.49,0.74}
\usepackage[pagebackref,breaklinks,colorlinks,allcolors=cvprblue]{hyperref}

\title{Quality-Aware Calibration for AI-Generated Image Detection in the Wild}

\author{Fabrizio Guillaro \ \ \ 
Vincenzo De Rosa \ \ \ 
Davide Cozzolino \ \ \ 
Luisa Verdoliva \\[2mm]
{University Federico II of Naples}\\[2mm]
}

\begin{document}

\twocolumn[{
\maketitle
\vspace*{-0.25in}
\centering
\includegraphics[width=0.98\linewidth, page=1, clip, trim=72 360 72 0]{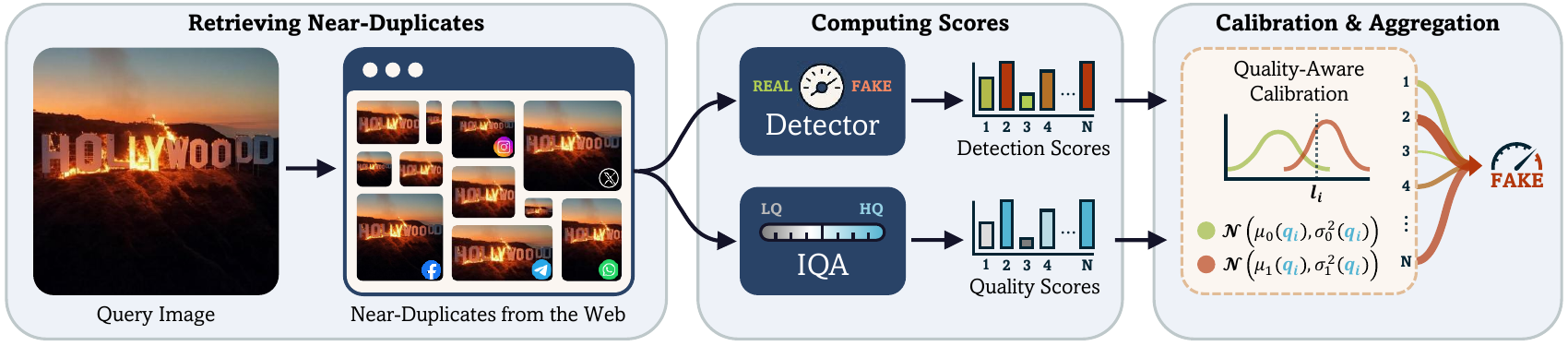}
\captionof{figure}{We study AI-generated image detection in real-world online settings.
Given a query image, we first retrieve near-duplicate versions from the web, which often differ in quality due to re-posting operations. We then run a forensic detector on each version to obtain detection scores, and an Image Quality Assessment (IQA) module to estimate its quality. Finally, we perform quality-aware calibration and aggregate the weighted scores across all the near-duplicate versions.}
\vspace{0.25in}
\label{fig:teaser}
}]

\begin{abstract} 
Significant progress has been made in detecting synthetic images, however most existing approaches operate on a single image instance and overlook a key characteristic of real-world dissemination: as viral images circulate on the web, multiple near-duplicate versions appear and lose quality due to repeated operations like recompression, resizing and cropping. As a consequence, the same image may yield inconsistent forensic predictions based on which version has been analyzed.
In this work, to address this issue we propose QuAD (Quality-Aware calibration with near-Duplicates) a novel framework that makes decisions based on all available near-duplicates of the same image.
Given a query, we retrieve its online near-duplicates and feed them to a detector: the resulting scores are then aggregated based on the estimated quality of the corresponding instance. By doing so, we take advantage of all pieces of information while accounting for the reduced reliability of images impaired by multiple processing steps.
To support large-scale evaluation, we introduce two datasets: AncesTree, an in-lab dataset of 136k images organized in stochastic degradation trees that simulate online reposting dynamics, and ReWIND, a real-world dataset of nearly 10k near-duplicate images collected from viral web content. Experiments on several state-of-the-art detectors show that our quality-aware fusion improves their performance consistently, with an average gain of around 8\% in terms of balanced accuracy compared to plain average. 
Our results highlight the importance of jointly processing all the images available online to achieve reliable detection of AI-generated content in real-world applications.
Code and data are publicly available at \url{https://grip-unina.github.io/QuAD/}.
\end{abstract}

\section{Introduction}
\label{sec:intro}

Generative AI has transformed how synthetic content is created, making high-quality media easy to produce even without technical expertise. Diffusion models accelerated this shift by improving quality and allowing users to create realistic scenes or alter content through simple prompts, tasks that once required significant expertise and cost \cite{Zhan2023multimodal}. While beneficial for creative and communication industries, these same capabilities enable fraud and disinformation at scale, creating major challenges for journalists, fact-checkers and law enforcement \cite{Barrett2024identifying, bontcheva2024generative}. 
Hence there is a strong request for detectors that can reliably distinguish real from AI-generated data \cite{Lin2024detecting}.

This is especially true over social networks where it is easy to spread disinformation. One main issue in this scenario is that images are often reshared multiple times, which causes the proliferation of several versions of them, more and more distorted.
Each reposting may introduce new artifacts and reduce the forensic traces thus making the detection much harder \cite{Li2025bridging, ricker2024AI-generated, porcile2023finding}. In fact, forensic detectors often rely on subtle statistical cues that can be strongly reduced by subsequent post-processing operations.
This causes the same detector to provide a very different scores on different versions of the same image and raises a fundamental question: which version should be trusted online? 

\begin{figure}[t!]
    \centering
    \includegraphics[width=0.95\linewidth, page=1,clip, trim=160 280 160 0]{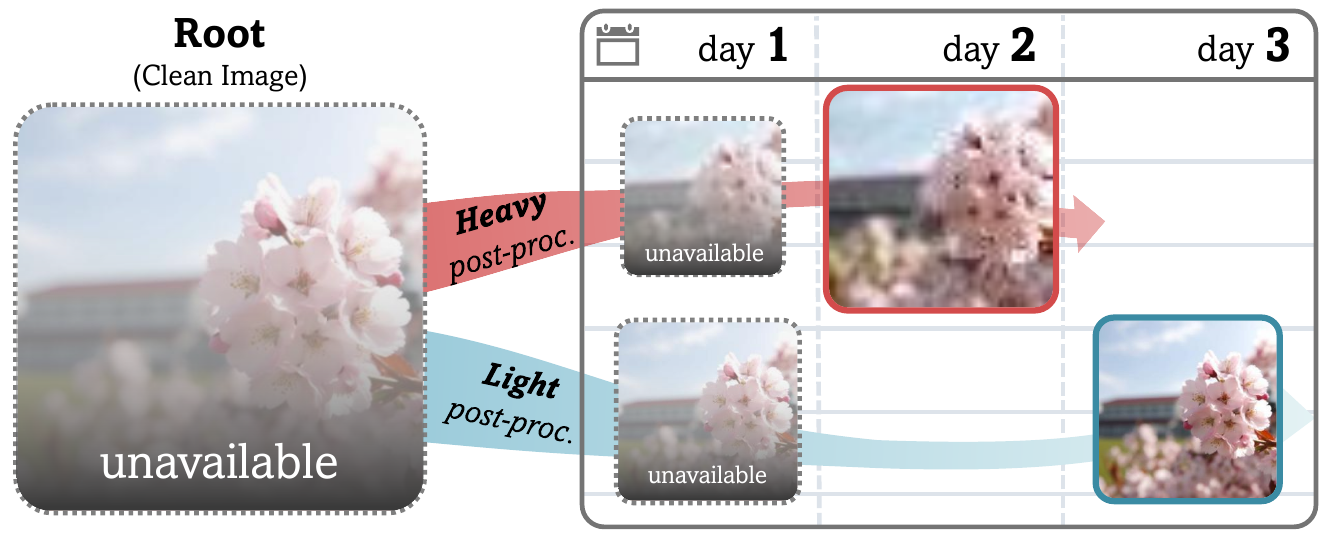}
    \caption{The oldest or largest image (day 2) is not necessarily the closest to the source in terms of quality, with some earlier instances possibly being unavailable.}
    \label{fig:images_over_time}
\end{figure}

\begin{figure}[t!]
    \centering
    \includegraphics[width=0.5\linewidth, page=2, clip, trim=-25 0 0 0]{figures/quality_midjourney.pdf}
    \includegraphics[width=1\linewidth, page=1]{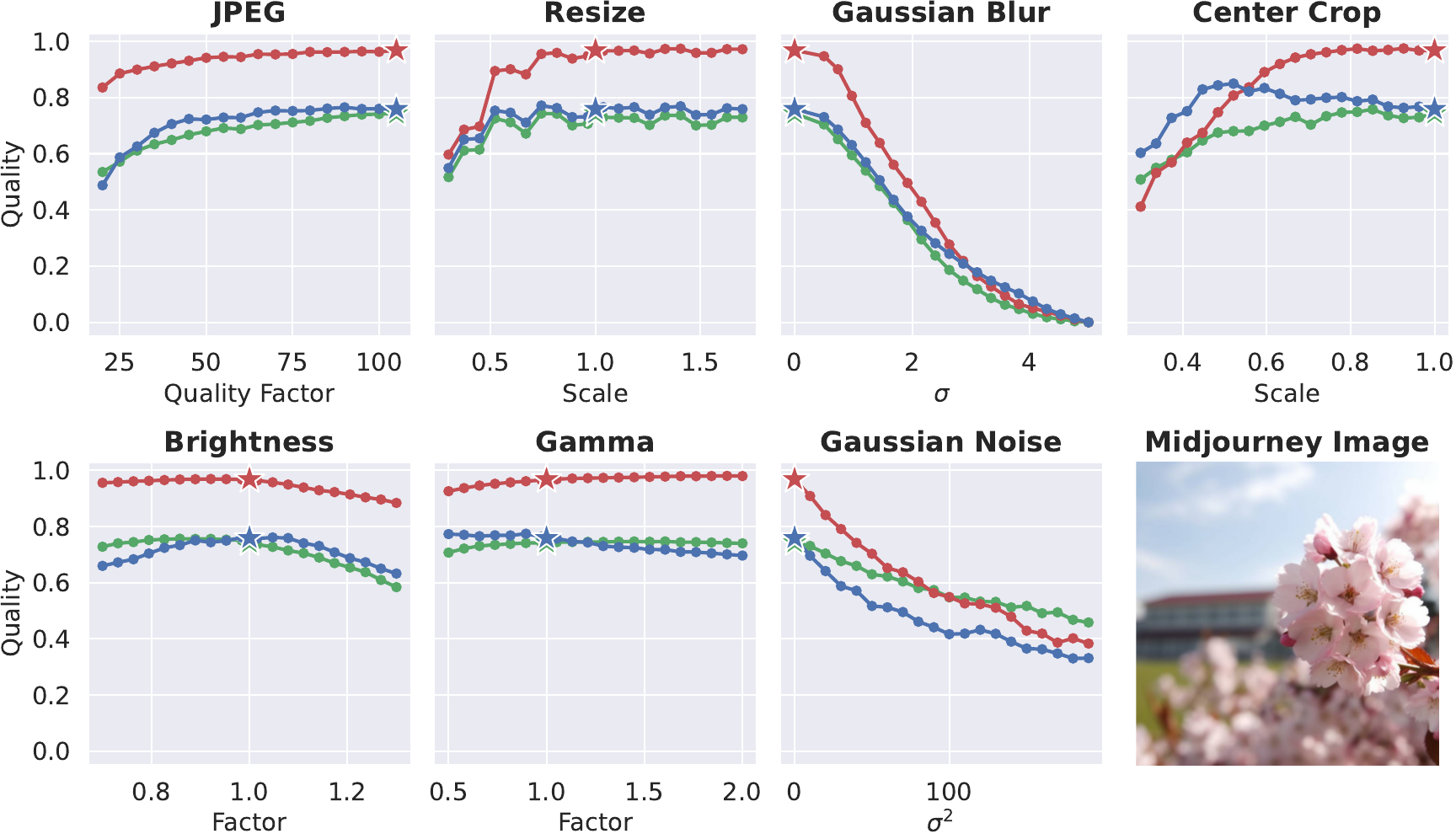}
    \caption{Impact of post-processing operations on the estimated quality of a synthetic image provided by TReS \cite{Golestaneh2022noreference}, QCN \cite{Shin2024blind}, LoDa \cite{Xu2024boosting} IQA methods. The star ($\star$) represents the clean image without degradation. The quality values have been normalized to facilitate comparisons across similar scales. We can easily observe that the quality gets worse as the degradations increase: the operations with most significant impact are compression, downsampling, blurring, and gaussian noise.}
    \label{fig:quality}
\end{figure}

In a recent study \cite{Karageogiou2024evolution}, it was concluded that examining the first instance of an image found online (presumably the one closest to the original upload) tends to yield more reliable detection results.
However, identifying the {\em original} or most reliable instance of an image in such a dynamic online environment is far from straightforward. The first or the largest version that appears online is not necessarily the true original, nor the one with the highest quality as its complete history is unknown and many ancestors may not be available anymore (Fig.~\ref{fig:images_over_time}). Moreover,
temporal metadata such as upload timestamps can be unreliable due to reposting delays, metadata manipulation, or discrepancies between platforms.
Simply aggregating information from all available versions does not necessarily lead to more accurate results, as the presence of heavily compressed or degraded copies can bias the outcome and increase uncertainty. This is further demonstrated by the analysis in Fig.~\ref{fig:quality}, which shows how the quality of an image, estimated with off-the-shelf Image Quality Assessment (IQA) methods, gets worse if an image is subjected to low-level processing operations, such as blurring, compression, resizing. Consequently, the problem extends beyond the detection of synthetic content in a single image: it involves reasoning across multiple versions of the same visual data to infer which one offers the most reliable content for the forensic analysis. 

In this work we aim at addressing this issue, which is overlooked in the current literature and poses challenges for journalists and fact-checkers everyday.
To the best of our knowledge, this is the first study that proposes a practical solution to this problem. 
Overall, we make the following contributions:
\begin{itemize}
    \item we propose \NAME\, (Quality-Aware calibration with near-Duplicates) a novel strategy that analyzes multiple versions of the same image (near-duplicates) and calibrate the corresponding detector scores based on their quality;
    \item we collect a real-world dataset (\NAMEvirali) of nearly 10k near-duplicate instances of images, downloaded from different web sources to capture authentic online degradations. This is the first large dataset of real and AI-generated images with near-duplicates versions retrieved online;
    \item we build an in-lab dataset (\NAMEinlab) of about 136k near-duplicates which mimics the real-world dataset but enables large-scale analysis under controlled conditions;
    \item we carry out an extensive evaluation of current SoTA detectors and show that our approach consistently improves their performance in terms of balanced accuracy.
\end{itemize}

\section{Related Work} 
\label{sec:related}

\paragraph{Robust forensic detectors.}
A number of forensic detectors have been proposed to cope with the degradations introduced by social-network. Typical sharing pipelines apply recompression and resizing, and may also include cropping and user-driven enhancements (e.g., filtering or contrast adjustment); all of which can attenuate or even remove the fine-grained forensic traces on which many detection methods rely \cite{Tariang2024synthetic}.
A simple and effective strategy to address such issue is {\em data augmentation} during training. In \cite{Wang2020cnn} it has been shown that JPEG compression and blur augmentation significantly improves robustness and generalization. A more intense augmentation would arguably guarantee further improvements \cite{Corvi2023detection} though it cannot incorporate all possible degradations paths that an image undergoes over a social network. 
Another difficulty is that current platforms do not disclose the operations applied on the shared data. Hence, in \cite{Wu2022robust} a noise modeling scheme is proposed that mimics predictable and unseen noise introduced by online social networks. Other methods rely on semantic high-level cues that turn out to be more resilient to low-level distortions \cite{Huang2025sida, Xia2025Mirage}, though such visual artifacts are very likely to disappear as generative models continue to improve.

All these approaches propose strategies to increase robustness of the detectors but do not address the problem of dealing with different online versions of the same image. The only work that analyzed this scenario is \cite{Karageogiou2024evolution}, where it is proposed to leverage a retrieval-assisted detection pipeline so as to select only the initial uploads that are less susceptible to post-processing operations. 
In this work we make one step further and propose a method that, instead of selecting a single {\em best} instance, jointly analyzes all retrieved near-duplicates by weighing each detector score according to the estimated quality of the corresponding image, obtaining a new calibrated score that improves reliability.

\begin{table}[t!]
    \centering
    \renewcommand{\arraystretch}{1.2} 
    \resizebox{1.0\linewidth}{!}{\small
    \begin{tabular}{r C{25mm} C{25mm}}
    \toprule
     \textbf{Source} & \textbf{\# Real/Fake sources} & \textbf{ \# Real/Fake near-duplicates} \\
    \midrule
    B-Free \cite{Guillaro2025bfree} & 20 / 15 & 1261 / 1851\\ 
    FOSID  \cite{Karageogiou2024evolution} & -~~/ 3  & ~~~-~~/ 517 \\ 
    AMMeBa \cite{Dufour2024ammeba} & 44 / 39 & 1929 / 2135 \\  
    Fact-Check Tool & ~-~~/ 18  & ~~~-~~/ 776 \\ 
    RRDataset \cite{Li2025bridging} & 23 / ~-~~~ &  1177 / ~~-~~~~~~~ \\ 
    \midrule
    Total & 87 / 75 &  4367 / 5279 \\ 
    \bottomrule
    \end{tabular}
    }
    \caption{Summary of the data sources used to build the ReWIND dataset, reporting for each source the number of real/fake original sources and the number of real/fake near-duplicate images collected online.
    }
    \label{tab:viral-dataset-sources}
\end{table}

\paragraph{Image quality assessment.} 

Image quality assessment (IQA) methods are designed to automatically predict the perceptual quality of an image, either by comparing it to a reference (FR-IQA) or by predicting quality without a reference (NR-IQA).
Quality analysis impacts the performance of forensic detectors as shown in \cite{kim2024correlation}, that demonstrates a significant variability in quality both among different deepfake datasets and within the data of each dataset. However only a few methods take into account the image quality analysis to build a forensic detector. 
Jiang et al. \cite{Jiang2025new} leverage image-quality assessment for deepfake discrimination, using IQA scores as discriminative features to separate real from fake faces, while Song et al. \cite{Song2024not} propose augmenting low-quality samples during training with a curriculum-learning strategy to improve realism. Our approach is different from such works, since we consider a different scenario and use IQA estimates at inference time to calibrate the detector score for each near-duplicate and to weigh its contribution in the final fusion. We would like to stress that in our framework we do not treat IQA scores as indicators of authenticity, and with the term quality we do not refer to the visual realism of the generated content, but rather to the level of image degradation.

\begin{figure}[t!]
    \centering
    \includegraphics[width=\linewidth, page=1]{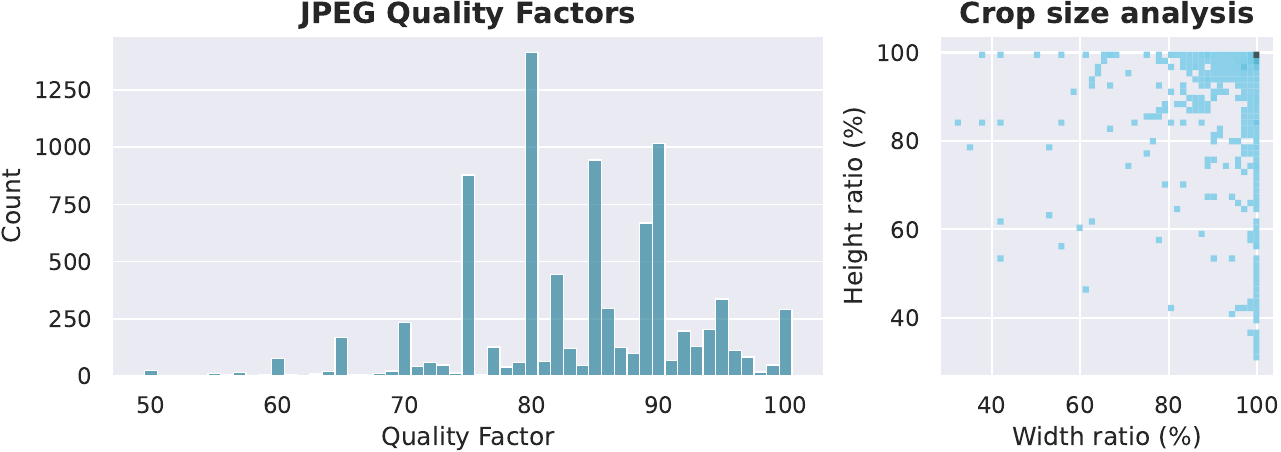}
    \caption{Distribution of JPEG quality factors (left), and crop size analysis (right) in \NAMEvirali.
    Left: histogram of extracted JPEG quality factors, showing that most images are compressed at high quality (roughly 70–95). Right: crop-size analysis, reporting the width and height ratios (in \%) relative to the reference image; most near-duplicates preserve a large portion of the original content (typically greater than 70-80\%).}
    \label{fig:viral_stats}
\end{figure}

\begin{figure*}
    \centering
    \includegraphics[width=1.0\linewidth, page=2,clip, trim=10 220 50 0]{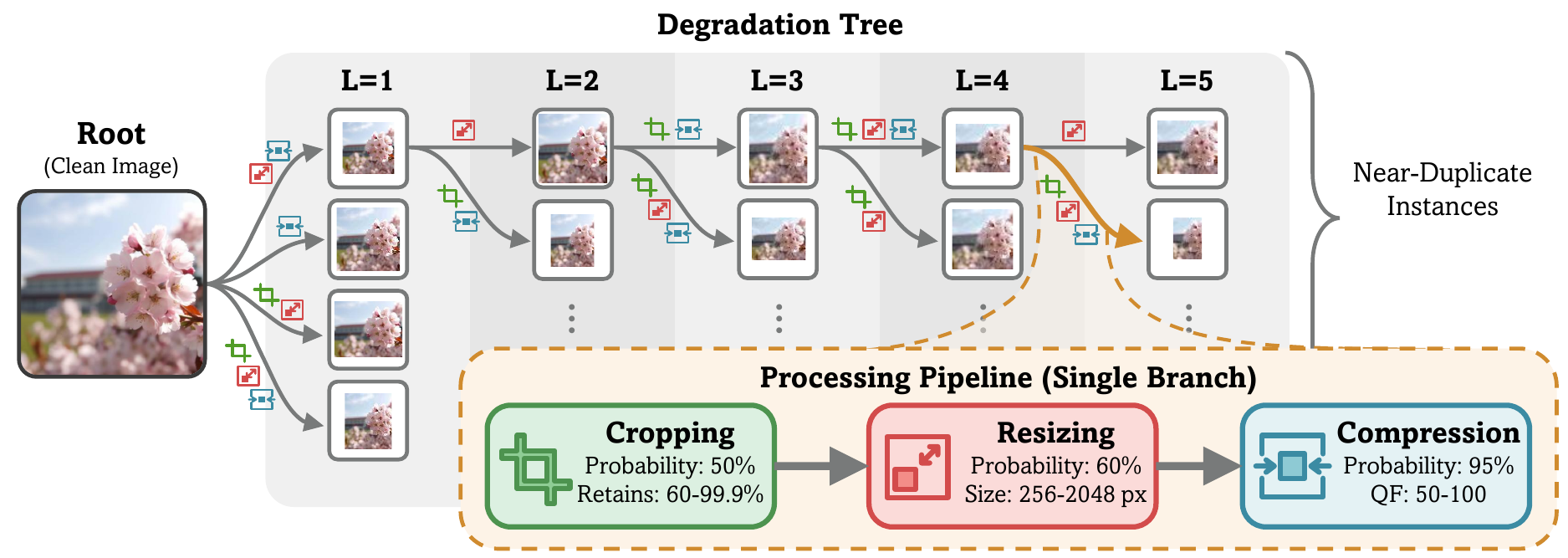}
    \caption{\NAMEinlab, we build a tree of progressive degradations used to generate near-duplicate image instances. Starting from a clean image, multiple degradation operations are applied across levels ($L=1$ to $L=5$). Each branch represents a sequential processing pipeline consisting of random cropping, resizing, and compression for a total of 124 near-duplicate samples for each image.
    }
    \label{fig:degradation_tree}
\end{figure*}

\section{Datasets}
\label{sec:datasets}

To simulate the variety of post-processing effects introduced when images are re-posted across the web, we first collected real-world examples of images that went viral on social networks. This allowed us to characterize the distribution of distortions observed online and to replicate it when constructing a more controlled in-lab dataset.
Below, we describe the two datasets used in our experiments: \NAMEvirali, a realistic dataset composed of images and multiple re-posted versions found over the web, and \NAMEinlab, a synthetic dataset generated by applying degradations that matches a realistic scenario.

\paragraph{\NAMEvirali: Real-World Images with Near-Duplicates.}

Several recent works \cite{Karageogiou2024evolution, Guillaro2025bfree, DellAnna2025truefake, Li2025bridging} released datasets containing images downloaded from the web. While \cite{Karageogiou2024evolution, Guillaro2025bfree} collected a few viral images with corresponding near-duplicates from the web, \cite{DellAnna2025truefake, Li2025bridging} created a dataset of images that have been manually uploaded on social networks by the authors. However, the latter studies include either single uploads \cite{DellAnna2025truefake} or consider only a few cross-platform repostings \cite{Li2025bridging}, without providing the intermediate instances.
For a better representation of realistic conditions, we collected a larger set of in-the-wild images that were actually shared on-line and became viral. The widespread circulation of these images allowed us to scrape the web to find multiple instances (i.e., near-duplicates) of each source image with different and unknown degradations.

We started with the set of viral images used in \cite{Guillaro2025bfree} and \cite{Karageogiou2024evolution}, and repeated the online search process to retrieve additional near-duplicate versions available on the web.
In addition, we integrated a set of fact-checked real and fake images from the AMMeBa fact-checking dataset \cite{Dufour2024ammeba}, a subset of real images by RRDataset \cite{Li2025bridging} coming from news websites, and also included some images from Google’s Fact Check Tools\footnote{\url{https://toolbox.google.com/factcheck/explorer/}} already confirmed to be AI-Generated. In order to find the near-duplicates for these additional sources, we leveraged Google’s Cloud Vision API\footnote{\url{https://cloud.google.com/vision}} .
A complete list of sources with the exact amount of images is summarized in Table~\ref{tab:viral-dataset-sources}.
We discarded small images whose shorter side was below 256 pixels and removed exact duplicates by filtering entries with identical MD5 checksums, which indicate the same file instance.
Finally, we only kept images with at least 10 near-duplicates, resulting in 162 sources (87 real / 75 fake), with a total of 9646 instances in different formats (JPEG, WebP and PNG) with the distribution of quality factors for the JPEG images shown in Fig.~\ref{fig:viral_stats}.

\paragraph{\NAMEinlab: In-lab dataset with degradation tree.}
To mimic the complex variety of real-world image degradation we designed the pipeline of operations described below.
For each source image, we generated a tree of depth $L=5$, represented in Fig.~\ref{fig:degradation_tree}, where branches represent random processing operations and nodes represent progressively degraded instances of the root (original clean image). To simulate high initial viral spread, the first level of the tree has four branches, with subsequent levels having a branching factor of two. Without the root, which is discarded as it represents the clean image before being uploaded, each tree yields a total of 124 near-duplicates for its source.
Each branch randomly applies a sequence of cropping, resizing, and compression operations.
More specifically, with 50\% probability, we apply a \textbf{random crop} preserving 60\%--99.9\% of the image area. Affine alignment of viral near-duplicates shows that roughly half are cropped, typically along a single axis (Fig.~\ref{fig:viral_stats}, right). We therefore crop one dimension at a time, which can yield 2D crops when applied repeatedly across successive instances. \textbf{Resizing} is applied with a probability of 60\% (so that the shorter side falls in the range [256-2048] pixels). 
To ensure enough variety, we randomly select between Pillow and OpenCV libraries and randomize the interpolation modality (bilinear, bicubic, or lanczos). Finally, \textbf{lossy compression} is performed with a chance of 95\%. The format (JPEG or WebP) and the Quality Factors (QFs) are sampled to match the real-world format proportions and quality distribution (Fig.~\ref{fig:viral_stats}). For JPEG compression, we randomly use two encoders (Pillow/OpenCV) to simulate the diversity of implementations found in the wild.

Real images come from the RAISE dataset \cite{nguyen2015raise}. 
For each real image, 10 synthetic images are generated using the caption of its real counterpart by SD1.4, SD2, SD-XL, Midjourney v5, DALL-E 2, DALL-E 3, Firefly (provided by \cite{Bammey2023synthbuster}), Flux and SD3 (provided by \cite{Guillaro2025bfree}) and Latent Diffusion. With this approach, real and fake images are aligned in terms of semantic content to mitigate possible biases.
We have a total of 100 real images and 1,000 synthetic images with 124 instances each, for a total of 136,400 images.

\begin{figure}[t!]
    \centering
    \includegraphics[height=0.04\linewidth, page=2, trim=-40 0 0 0]{figures/synth_levels.pdf}
    \includegraphics[width=1\linewidth, page=1, trim=0 0 0 -5]{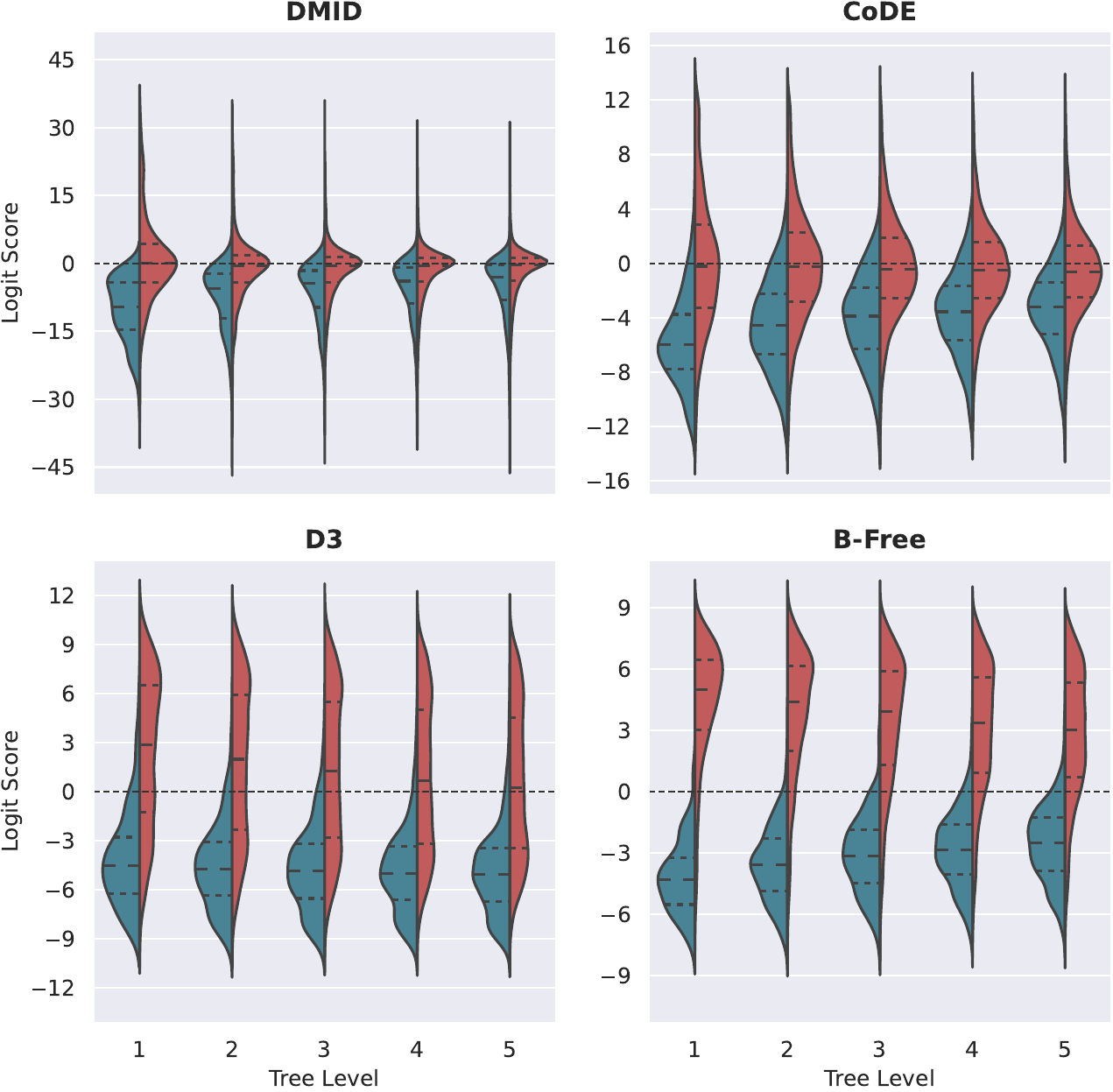}
    \caption{Score distributions of several forensic detectors (DMID~\cite{Corvi2023detection}, CoDE~\cite{Baraldi2024contrastive}, D3~\cite{Yang2025d3}, and B-Free~\cite{Guillaro2025bfree}) for real and fake images at different levels of the tree. The higher the level of degradation, the more the distributions overlap, causing confusion between real and fake.}
    \label{fig:hists_level}
\end{figure}

\section{Proposed Method}
\label{sec:method}

In real-world online scenarios, the same image often circulates in multiple near-duplicate versions, each undergoing different post-processing. As a result, the same forensic detector may produce significantly different scores depending on which version is analyzed. To address this issue, we propose \NAME, a quality-aware calibration approach. We first estimate the quality of each retrieved instance using a no-reference image quality assessment metric, and then calibrate the detector's score as a function of image quality. Specifically, we model the distribution of detector logits conditioned on quality, separately for real and fake images, and use this information to weigh each copy’s contribution in the final decision. In this way, low-quality versions, where real and fake distributions tend to overlap more, have a reduced impact, while more reliable instances guide the final fusion. In the following we describe our approach.

\paragraph{Background.}
In our scenario, we have $N$ near-duplicate instances ($X_1, X_2, \ldots, X_N$) that correspond to the same original image.
For each instance $X_i$, the detector outputs a logit score $l_i$ that estimates the logarithm of the a posteriori likelihood ratio. If we define a binary label $y \in {0,1}$, where $y=0$ denotes a real image and $y=1$ a synthetic one, then:

\begin{equation}
l_i = \log \frac{P(y=1 \mid X_i)}{P(y=0 \mid X_i)}
\end{equation}
Given the availability of $N$ instances, the MAP decision rule for classifying the image as fake is:

\begin{equation}
\log \frac{P(y=1 \mid X_1, X_2, \ldots, X_N)}{P(y=0 \mid X_1, X_2, \ldots, X_N)} > 0
\end{equation}
We adopt an approximation
assuming equal prior probabilities for real and fake images, and conditional independence of the instances given the label. Even if this is not satisfied in practice, it makes the problem easier to handle. The formula hence reduces to:

\begin{figure}[t!]
    \centering
    \includegraphics[height=0.04\linewidth, page=2, trim=-40 0 0 0]{figures/synth_levels.pdf}
    \includegraphics[height=0.04\linewidth, page=2, trim=0 0 0 0]{figures/scatters.pdf}
    \includegraphics[width=0.99\linewidth, page=1, trim=0 0 0 -5]{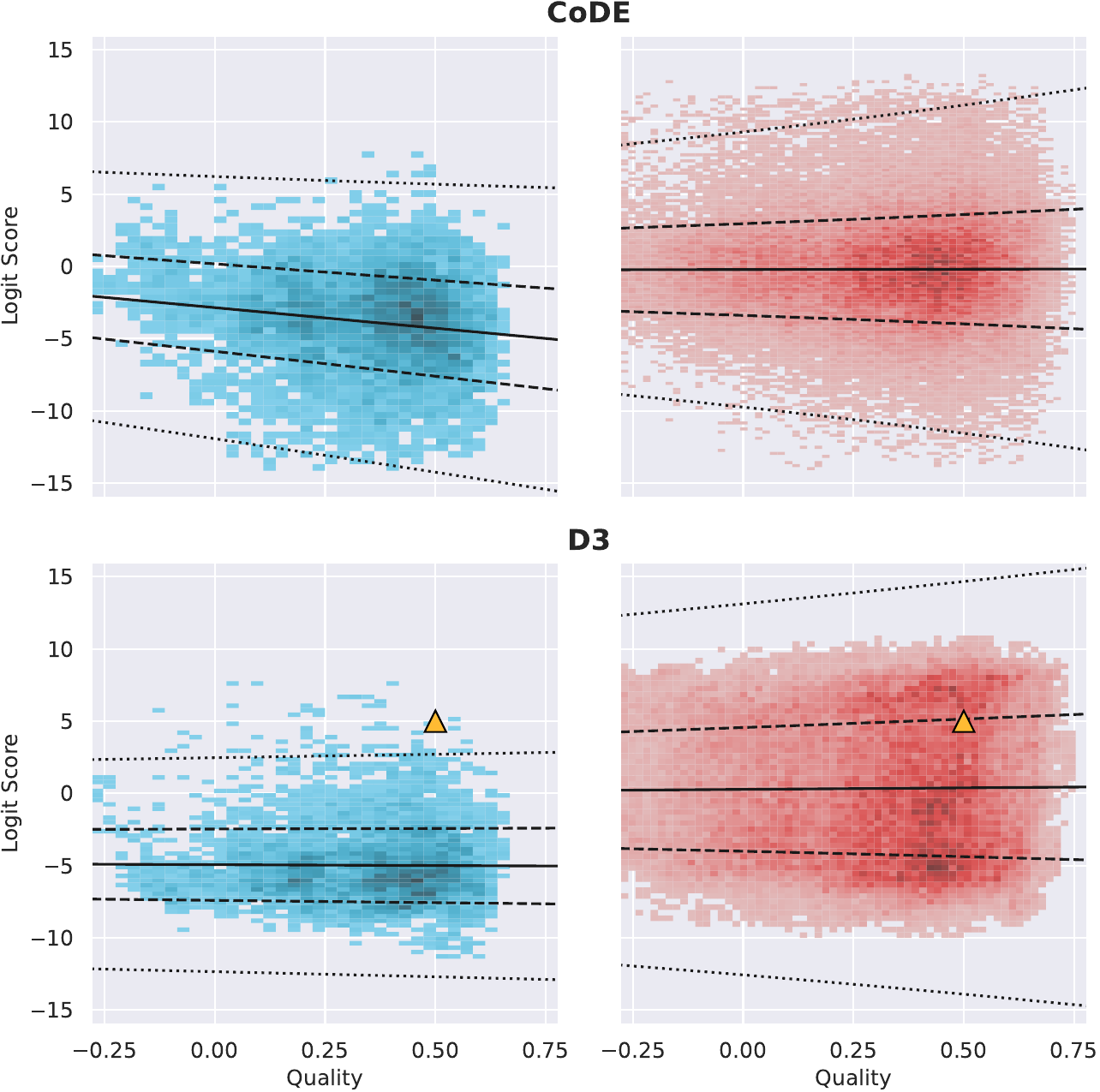}
    \caption{Distribution of the scores of two detectors (CoDE~\cite{Baraldi2024contrastive}, and D3~\cite{Yang2025d3}) as a function of image quality for real (left) and fake (right) images. 
    Black lines represent the Gaussian fitting; solid lines indicate the mean trend, while dashed and dotted lines show the spread at $\pm1$ and $\pm3$  standard deviations, respectively.}
    \label{fig:fitting}
\end{figure}

\begin{equation}
\sum_{i=1}^{N} \log \frac{P(X_i \mid y=1)}{P(X_i \mid y=0)}
= \sum_{i=1}^{N} l_i > 0 .
\end{equation}

This straightforward Bayesian approach combines evidence from multiple instances by aggregating their logits under the conditional independence assumption.
However, such naïve fusion can be unreliable, since progressive degradations reduce class separability and make some predictions less informative \cite{Karageogiou2024evolution}.
This effect is also visible in Fig.~\ref{fig:hists_level}, where the logit score histograms for real and fake images increasingly overlap as the number of processing operations grows (i.e. increasing depth in the tree). These score distributions are shown for different forensic detectors. While their specific shapes vary with the performance of each detector, they exhibit a consistent trend.

\paragraph{Quality-aware calibration.}
Now the main question to answer is how to understand which scores are reliable and which ones are not.
In a well-calibrated system, the answer is straightforward: the absolute value of the logit score also reflects the reliability of the prediction. Therefore, the key is to calibrate the model. Our idea is to guide this calibration using the image quality.
In this way, less reliable logit scores should have a smaller impact on the overall sum, while more reliable logit scores should contribute more significantly.
To obtain this effect, we propose predicting the degree of degradation using a no-reference image quality index. Indeed, Fig.~\ref{fig:quality} shows that a no-reference quality index is correlated with the strength of the post-processing operations applied to the image.
A straightforward solution would be to consider only the images with the highest quality index, thereby limiting the use of samples. However, this would discard potentially useful information. Our approach instead relies on using a Gaussian fitting as described below.

\paragraph{Gaussian fitting.}
We model the behavior of the logit score $l_i$ as a function of the no-reference quality index $q_i$ using a Gaussian fitting separately for real and fake images:

\begin{equation}
\begin{aligned}
    l_i \mid q_i,y=1 &\sim \mathcal{N}\left(\mu_1(q_i), \sigma_1^2(q_i) \right) \\
    l_i \mid q_i,y=0 &\sim \mathcal{N}\left(\mu_0(q_i), \sigma_0^2(q_i) \right)
\end{aligned}
\end{equation}
Note that both the mean and variance depend on $q_i$.
In this way, the resulting decision rule becomes:

\begin{equation}
    \sum_{i=1}^N \log \frac{P(l_i  \mid q_i,y=1)}{P(l_i \mid q_i,y=0)} = \sum_{i=1}^N \hat{l}_i  > 0 
\end{equation}
where $\hat{l}_i$ denotes the corrected logit score, defined as:
\begin{equation}
\hat{l}_i = \frac{(l_i - \mu_0(q_i))^2}{2\sigma_0^2(q_i)} -\frac{(l_i - \mu_1(q_i))^2}{2\sigma_1^2(q_i)} +  \log\left(\frac{\sigma_o(q_i)}{\sigma_1(q_i)}\right)
\end{equation}
As a result, the absolute value of the corrected logit score depends on the degree of overlap between the Gaussian distributions.
For low-quality instances, Gaussians are expected to be closer and overlap more, resulting in a corrected logit score closer to zero and consequently reducing its contribution to the final sum.
For both statistics we assume a simple linear dependence with $q_i$:
\begin{equation}
\begin{aligned}
    \mu_j(q_i) &= a_j \cdot q_i + b_j \;\;\;\;\;\; j=0,1\\
    \log \sigma_j^2(q_i) &= \alpha_j \cdot q_i + \beta_j  \;\;\;\;\;\; j=0,1\\
\end{aligned}
\end{equation}
where the eight coefficients ($a_0, b_0, \alpha_0, \beta_0, a_1, b_1, \alpha_1, \beta_1$) are estimated using a Maximum Likelihood strategy on a development set taken from around 50\% of the \NAMEinlab dataset (the remaining data are used for evaluation).

\paragraph{Score distributions analysis.}

Fig.~\ref{fig:fitting} illustrates the joint distribution of logit scores (y-axis) and no-reference quality indexes LoDa (x-axis) for real (left) and fake (right) images. We observe that the mean and the spread of the logit scores are influenced by the quality index. The fitting results are shown by the black lines, which represent the Gaussian distributions as a function of image quality.
At inference time, a test image (denoted by the orange marker in Fig. \ref{fig:fitting}), which is characterized by a logit score and an estimated quality, will be evaluated by comparing the real and the fake Gaussian distributions in that point.

\begin{figure}[t!]
    \centering
    \includegraphics[width=1\linewidth, page=1]{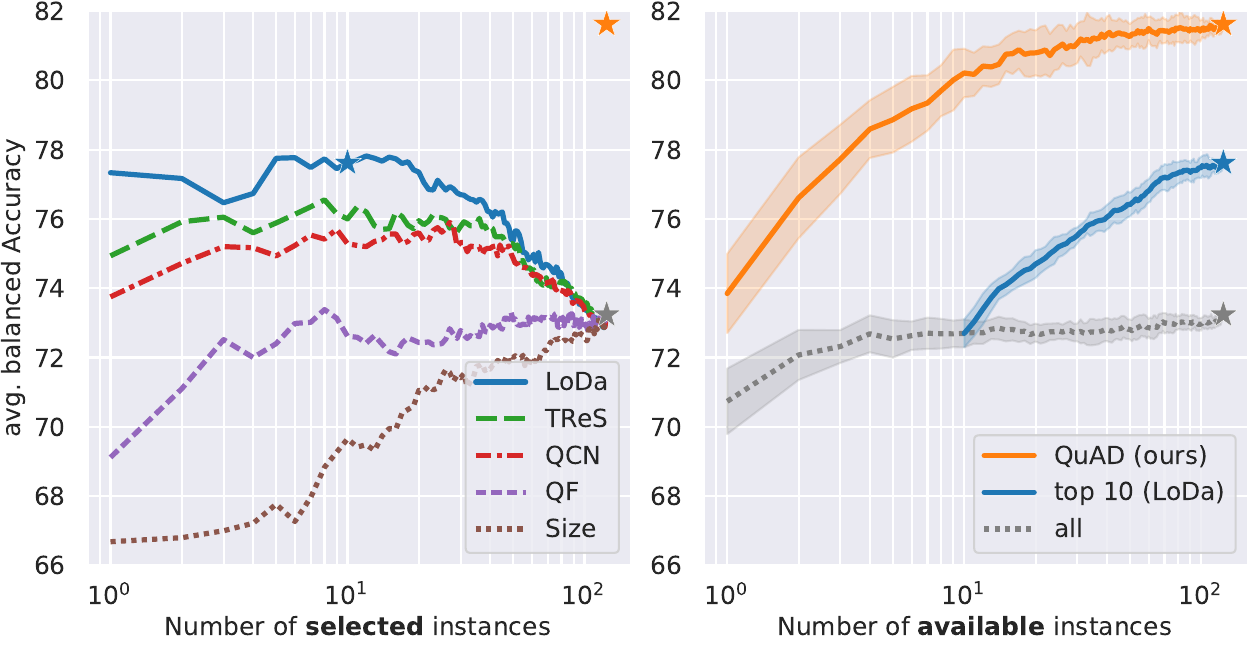}
    \caption{Performance in terms of average Balanced Accuracy across the six detectors on the \NAMEinlab~dataset. Left: the effect of different ranking by quality factor, image size, and IQA score (TReS \cite{Golestaneh2022noreference}, QCN \cite{Shin2024blind}, LoDa \cite{Xu2024boosting}). The orange $\star$ representing the accuracy of our proposed calibrated aggregation of all 124 instances. Right: performance when retrieving only a limited number of tree instances. The proposed \NAME~(orange) is compared with the combination of the available original logits (gray) and with the aggregation of the top 10 instances according to the LoDA index (blue).}
    \label{fig:acc_synthbuster_comparison}
\end{figure}

\begin{table*}[t!]
    \centering
	\small
	\scalebox{1.0}{
    \setlength{\tabcolsep}{6pt}
    \begin{tabular}{ll C{17mm}C{17mm}C{17mm}C{17mm}C{17mm}C{17mm} C{17mm}}
    \toprule
\multicolumn{2}{c}{\textbf{bAcc} $\uparrow$/ \textbf{NLL}$\downarrow$} &  \textbf{DMID} & \textbf{CoDE} & \textbf{D3} & \textbf{B-Free} & \textbf{DRCT} & \textbf{CO-SPY} & \textbf{AVG} \\
\cmidrule(r){1-2} \cmidrule(lr){3-8} \cmidrule(l){9-9}
oracle  & L1 & 75.5 / 1.24 & 73.9 / 0.85 & 79.7 / 0.59 & 97.3 / 0.09 & 68.6 / 0.59 & 77.6 / 0.60 & 78.8 / 0.66 \\
\cmidrule(r){1-2} \cmidrule(lr){3-8} \cmidrule(l){9-9}
random  &  1 & 63.3 / 1.68 & 65.9 / 1.01 & 74.5 / 0.92 & 88.4 / 0.32 & 57.9 / 1.10 & 74.4 / 0.77 & 70.7 / 0.97 \\
naive   &all & 62.1 / 1.17 & 71.8 / 0.72 & 77.4 / 0.77 & 94.3 / 0.16 & 57.9 / 0.98 & 75.9 / 0.70 & 73.2 / 0.75 \\
\cmidrule(r){1-2} \cmidrule(lr){3-8} \cmidrule(l){9-9}
        &  1 & 61.6 / 1.16 & 65.7 / 0.87 & 75.2 / 0.82 & 85.9 / 0.33 & 50.5 / 1.21 & 75.8 / 0.70 & 69.1 / 0.85 \\
QF      & 10 & 65.0 / 1.39 & 70.7 / 0.76 & 77.6 / 0.70 & 93.5 / 0.16 & 52.9 / 1.04 & 76.0 / 0.66 & 72.6 / 0.78 \\
        & 20 & 62.5 / 1.40 & 69.9 / 0.78 & 79.7 / 0.69 & 94.3 / 0.15 & 51.9 / 1.07 & 76.3 / 0.68 & 72.4 / 0.79 \\
\cmidrule(r){1-2} \cmidrule(lr){3-8} \cmidrule(l){9-9}
        &  1 & 58.1 / 1.12 & 57.2 / 1.02 & 72.2 / 0.94 & 84.2 / 0.40 & 55.9 / 1.41 & 72.5 / 0.78 & 66.7 / 0.95 \\
Size    & 10 & 60.7 / 0.81 & 58.7 / 0.87 & 77.0 / 0.72 & 93.0 / 0.27 & 53.0 / 1.26 & 75.5 / 0.65 & 69.6 / 0.76 \\
        & 20 & 60.8 / 0.76 & 61.1 / 0.83 & 79.8 / 0.70 & 95.7 / 0.23 & 53.0 / 1.24 & 75.9 / 0.65 & 71.1 / 0.73 \\
\cmidrule(r){1-2} \cmidrule(lr){3-8} \cmidrule(l){9-9}
     &  1 & 74.3 / 1.48 & 69.2 / 1.06 & 79.3 / 0.69 & 96.1 / 0.10 & 67.9 / 0.85 & 77.2 / 0.63 & 77.3 / 0.80 \\
LoDa & 10 & 73.3 / 1.09 & 75.5 / 0.74 & 79.9 / 0.64 & 98.0 / 0.08 & 62.2 / 0.83 & 76.8 / 0.62 & 77.6 / 0.66 \\
     & 20 & 72.0 / 1.04 & 76.2 / 0.70 & 79.4 / 0.67 & 98.4 / 0.08 & 60.7 / 0.85 & 77.4 / 0.62 & 77.4 / 0.66 \\
\cmidrule(r){1-2} \cmidrule(lr){3-8} \cmidrule(l){9-9}
\NAME & all   & 79.6 / 0.61 & 79.6 / 0.51 & 81.2 / 0.41 & 97.1 / 0.14 & 70.2 / 0.57 & 82.0 / 0.34 & 81.6 / 0.43 \\
        \bottomrule
    \end{tabular}
    }
    \caption{Performance on the \NAMEinlab~dataset in terms of balanced Accuracy and NLL. The second column denotes the number of instances per image considered in the aggregation, with "L1" referring to the oracle performance obtained by considering only the images at the first level of the tree (higher quality). Specifically, we show results for simple aggregation of $N$ near-duplicates per image according to different ranking strategies: random, compression quality, dimension and image quality (LoDa). The last row, \NAME, is our proposed calibrated aggregation.}
    \label{tab:accuracy_inlab}
\end{table*}

\section{Experimental Results}
\label{sec:results}

\paragraph{Metrics, Baselines and SoTA.}
We measure the performance by means of balanced Accuracy and the Negative Log-Likelihood (NLL) \cite{Quinonero2006evaluating}, that computes the similarity between the distribution of the model's predictions and the actual data distribution, penalizing both low confidence in the correct class and overconfidence in incorrect ones. 

We compare our approach with some baselines. First, we consider analyzing a single image ({\bf random}), computing an averaging score over the predictions on all the near-duplicate images ({\bf naive}) and only the instances at the first level ($L1$) of the tree. The latter is an ideal scenario ({\bf oracle}), since we do not usually have this information in advance.
Then for each realistic case we compute performance by simple aggregation (via logit score mean) of the top 1, 10, and 20 near-duplicates per image. These near-duplicates are ranked according to different strategies: the last known compression quality factor (which is available given our controlled in-lab setup), the overall image size, and the off-the-shelf IQA method LoDa \cite{Xu2024boosting}.
We apply these baselines and our strategy to several SoTA AI-generated image detectors whose code and pre-trained models are publicly available, briefly described in the following.

\bb{DMID \cite{Corvi2023detection}} is based on a ResNet-50 network with a critical architectural modification: the down-sampling at the first layer is removed to preserve low-level forensic artifacts. \bb{CoDE \cite{Baraldi2024contrastive}} instead leverages a Vision Transformer trained with a contrastive learning objective; in our evaluation, we report results from its linear probing configuration. \bb{D3 \cite{Yang2025d3}} learns shared artifacts between original and patch-shuffled images to capture more generalizable cues. \bb{B-Free \cite{Guillaro2025bfree}} addresses training bias by reconstructing synthetic images aligned with their authentic counterparts, ensuring semantic alignment. \bb{DRCT \cite{Chen2024drct}} combines diffusion-based reconstructions with contrastive learning; we report results using its ViT-based configuration trained on SD2 reconstructions. Finally, \bb{CO-SPY \cite{Cheng2025cospy}} integrates semantic features, extracted by a frozen CLIP, and artifacts features via VAEs from Stable Diffusion and a ResNet-50. 
These features are adaptively fused before the final classification.

\begin{table*}[t!]
    \centering
	\small
	\scalebox{1.0}{
    \setlength{\tabcolsep}{6pt}
    \begin{tabular}{ll C{17mm}C{17mm}C{17mm}C{17mm}C{17mm}C{17mm} C{17mm}}
    \toprule
\multicolumn{2}{c}{\textbf{bAcc}$\uparrow$ / \textbf{NLL}$\downarrow$} & \textbf{DMID} & \textbf{CoDE} & \textbf{D3} & \textbf{B-Free} & \textbf{DRCT} & \textbf{CO-SPY} & \textbf{AVG} \\
\cmidrule(r){1-2} \cmidrule(lr){3-8} \cmidrule(l){9-9}
random &  1 & 55.9 / 2.77 & 57.2 / 1.72 & 66.4 / 1.39 & 78.4 / 0.63 & 57.8 / 1.18 & 64.4 / 0.92 & 63.4 / 1.44 \\
naive  &all & 50.0 / 2.58 & 60.8 / 1.43 & 67.4 / 1.24 & 80.1 / 0.48 & 53.8 / 1.06 & 65.7 / 0.81 & 63.0 / 1.27 \\
\cmidrule(r){1-2} \cmidrule(lr){3-8} \cmidrule(l){9-9}
     &  1 & 62.9 / 2.16 & 56.2 / 1.62 & 69.1 / 1.27 & 79.6 / 0.53 & 70.9 / 1.23 & 65.3 / 0.95 & 67.3 / 1.29 \\
Date & 10 & 51.3 / 2.09 & 60.4 / 1.39 & 68.9 / 1.27 & 79.4 / 0.47 & 73.0 / 0.99 & 70.6 / 0.82 & 67.3 / 1.17 \\
     & 20 & 50.7 / 2.22 & 60.9 / 1.38 & 67.6 / 1.26 & 79.4 / 0.46 & 73.0 / 1.01 & 68.7 / 0.84 & 66.7 / 1.20 \\
\cmidrule(r){1-2} \cmidrule(lr){3-8} \cmidrule(l){9-9}
     &  1 & 52.9 / 5.29 & 60.3 / 1.55 & 71.2 / 1.23 & 81.2 / 0.48 & 55.2 / 2.00 & 61.6 / 0.89 & 63.7 / 1.91 \\
QF   & 10 & 50.0 / 4.26 & 58.8 / 1.46 & 67.4 / 1.21 & 81.4 / 0.46 & 52.7 / 1.36 & 68.4 / 0.80 & 63.1 / 1.59 \\
     & 20 & 50.0 / 3.71 & 60.8 / 1.46 & 68.7 / 1.21 & 80.1 / 0.48 & 53.4 / 1.22 & 65.7 / 0.79 & 63.1 / 1.48 \\
\cmidrule(r){1-2} \cmidrule(lr){3-8} \cmidrule(l){9-9}
     &  1 & 62.9 / 2.16 & 56.2 / 1.62 & 69.1 / 1.27 & 79.6 / 0.53 & 59.8 / 1.13 & 65.3 / 0.95 & 65.5 / 1.28 \\
Size & 10 & 51.3 / 2.09 & 60.4 / 1.39 & 68.9 / 1.27 & 79.4 / 0.47 & 52.3 / 1.07 & 70.6 / 0.82 & 63.8 / 1.19 \\
     & 20 & 50.7 / 2.22 & 60.9 / 1.38 & 67.6 / 1.26 & 79.4 / 0.46 & 53.1 / 1.08 & 68.7 / 0.84 & 63.4 / 1.21 \\
\cmidrule(r){1-2} \cmidrule(lr){3-8} \cmidrule(l){9-9}
     &  1 & 64.1 / 1.62 & 59.0 / 1.71 & 70.4 / 1.30 & 84.9 / 0.41 & 57.2 / 1.17 & 60.7 / 1.07 & 66.1 / 1.21 \\
LoDa & 10 & 56.7 / 2.01 & 64.9 / 1.26 & 69.7 / 1.20 & 85.4 / 0.39 & 56.6 / 0.98 & 62.7 / 0.95 & 66.0 / 1.13 \\
     & 20 & 53.3 / 2.29 & 62.2 / 1.29 & 68.7 / 1.22 & 84.8 / 0.42 & 55.6 / 0.99 & 65.0 / 0.91 & 64.9 / 1.19 \\
\cmidrule(r){1-2} \cmidrule(lr){3-8} \cmidrule(l){9-9}
\NAME  & all & 66.7 / 0.64 & 64.0 / 0.63 & 72.0 / 0.77 & 86.1 / 0.35 & 70.5 / 0.58 & 62.3 / 0.83 & 70.3 / 0.63 \\
\NAME* & all & 61.0 / 0.64 & 70.0 / 0.59 & 71.5 / 0.61 & 89.6 / 0.32 & 70.9 / 0.60 & 65.4 / 0.62 & 71.4 / 0.57 \\
\bottomrule
    \end{tabular}
}
\caption{Performance on the \NAMEvirali~dataset of real-world viral images in terms of balanced Accuracy and NLL. The second column denotes the number of instances per image considered in the aggregation. We report results for simple aggregation of $N$ near-duplicates per image according to different ranking strategies: random, date, compression quality, dimension and image quality (LoDa). The last two rows present two variants of our method. For the second last one (\NAME), the coefficients are estimated on \NAMEinlab~development set, while for the last one (\NAME*), the coefficients are estimated on the same \NAMEvirali~dataset using a leave-one-out strategy.
}
    \label{tab:accuracy_viral}
\end{table*}

\vspace{-1mm}
\paragraph{Analysis on \NAMEinlab.}
Table~\ref{tab:accuracy_inlab} reports results on the \NAMEinlab~ dataset that clearly shows that aggregating multiple scores on images ranked by their quality is beneficial. For example, considering top 10 instances ranked by LoDa (77.6\% bAcc, 0.66 NLL) is better than considering solely the best LoDa-ranked instance (77.3\% bAcc, 0.80 NLL) and the simple aggregation of all instances (73.2\% bAcc, 0.75 NLL). Of course, analyzing a single randomly selected web image does not yield reliable results (70.7\% bAcc, 0.97 NLL), which is what a fact-checker would typically do using an off-the-shelf forensic detector.
However, leveraging all instances and calibrating the logits prior to the aggregation achieves the best average performance (81.6\% bAcc, 0.43 NLL), as this strategy ensures that lower quality instances have less impact on final score.
The proposed strategy does not appear to be sensitive to the choice of IQA score. Indeed, when the other IQA scores, TReS \cite{Golestaneh2022noreference} and QCN \cite{Shin2024blind}, are used for calibration, the average accuracies are 81.5\% and 81.4\%, respectively.
Similar results are also obtained when using a second-order model.

Fig.~\ref{fig:acc_synthbuster_comparison} (left)
shows the effect of ranking by quality factor, image size, and IQA score (TReS \cite{Golestaneh2022noreference}, QCN \cite{Shin2024blind}, LoDa \cite{Xu2024boosting})
in terms of average balanced accuracy on the six detectors, with our approach of calibrating and aggregating all 124 instances being represented by the orange $\star$.
It is clear that the compression quality factor and especially the image dimensions are not reliable as a ranking metric. In fact, there is no guarantee that the largest image has not been heavily processed before a final upscaling. We can also observe that the best IQA approach is LoDa that achieves the highest balanced accuracy.
All the ranking curves converge to the same point (gray star), which coincides with the naive aggregation of all 124 instances.
The right panel of Fig.~\ref{fig:acc_synthbuster_comparison} instead reports performance in the setting where not all near-duplicates are available.
The gray curve corresponds to a simple aggregation of all the available instances, while the blue curve aggregates only the 10 best instances according to the LoDA index.
The orange curve corresponds to the proposed aggregation based on corrected logit scores.
This figure also accounts for a scenario were near-duplicates are rare. Even in this situation the proposed aggregation works well with the few examples available.

\vspace{-1mm}
\paragraph{Analysis on \NAMEvirali.}
Given the unconstrained nature of viral content, we extend our evaluation on the \NAMEvirali~dataset to assess performance in a much more challenging real-world scenario.
Unlike the controlled in-lab setting, the degradation history of each instance is much more complex and completely unknown.
In this situation, many earlier versions may not be accessible anymore online, and the diffusion process is not limited to the five repostings that we simulated in the controlled dataset, with a general performance degradation.
In Table~\ref{tab:accuracy_viral} we present the balanced accuracy and NLL performance on \NAMEvirali.
In this Table we introduce an additional ranking strategy based on the upload date, as proposed in \cite{Karageogiou2024evolution}.
As previously discussed, relying on the date and processing one single image leads on average to 67.3\% of Accuracy and 1.29 of NLL, while our approach, estimating the coefficients on \NAMEinlab~development set, obtains 70.3\% of Accuracy and 0.63 of NLL.

The proposed solution achieves a significant performance improvement over the simple aggregation of all instances for almost all detectors.
Only the CO-SPY detectors exhibit a marginal reduction, probably the \NAMEinlab~development set is not sufficient to capture all the variability of real-world scenarios.
Indeed, estimating the coefficients on the same \NAMEvirali~dataset using a leave-one-out strategy, the average performance has a further increase (71.4\% bAcc, 0.57 NLL).
The performance on ReWIND suggests that the approach generalizes well, despite being fitted on the small set of sources of AncesTree. This is due to the fact that fake images are semantically aligned with real images. As a result, the fitting process focuses on the generation artifacts rather than on image content.

\section{Conclusions}
\label{sec:conclusions}

In this paper we address the problem of detecting AI-generated images in the wild, where multiple copies of the same image are present online. In fact, once an image is posted, it is can be copied, resized, re-encoded, and reshared, producing many versions that are similar but not identical. The problem is that forensic detectors can produce very different scores depending on which processed version is evaluated. Prior work recommends relying on earlier online versions that presumably underwent less processing, unfortunately such images can be difficult to retrieve. 
To this end, we propose \NAME, a method that combines calibrated scores of near-duplicate versions based on the image quality. Experiments show that our approach outperforms baselines that rely on the earliest posted image or that select versions by resolution or quality.

In future work we will further explore the retrieval phase, analyzing how errors related to missed near-duplicates or false positives affect the overall performance, and investigate strategies to filter out irrelevant samples. 
We will also consider adversarial scenarios, where near-duplicates could have been intentionally manipulated to mislead the detector and finally will extend such analysis to fully AI-generated videos.

\vspace{2mm}
\noindent
\bb{Acknowledgments.} 
We gratefully acknowledge the support of this research by a Google Gift. We would also like to thank Avneesh Sud and Ben Usman for useful discussions.

\balance 
{
    \small
    \bibliographystyle{ieeenat_fullname}
    \bibliography{main}

\begin{thebibliography}{30}
\providecommand{\natexlab}[1]{#1}
\providecommand{\url}[1]{\texttt{#1}}
\expandafter\ifx\csname urlstyle\endcsname\relax
  \providecommand{\doi}[1]{doi: #1}\else
  \providecommand{\doi}{doi: \begingroup \urlstyle{rm}\Url}\fi

\bibitem[Bammey(2023)]{Bammey2023synthbuster}
Quentin Bammey.
\newblock {Synthbuster: Towards detection of diffusion model generated images}.
\newblock \emph{IEEE Open Journal of Signal Processing}, 2023.

\bibitem[Baraldi et~al.(2024)Baraldi, Cocchi, Cornia, Baraldi, Nicolosi, and Cucchiara]{Baraldi2024contrastive}
Lorenzo Baraldi, Federico Cocchi, Marcella Cornia, Lorenzo Baraldi, Alessandro Nicolosi, and Rita Cucchiara.
\newblock {Contrasting Deepfakes Diffusion via Contrastive Learning and Global-Local Similarities}.
\newblock In \emph{ECCV}, 2024.

\bibitem[Barrett et~al.(2024)Barrett, Boyd, Burzstein, Carlini, Chen, Choi, Chowdhury, Christodorescu, Datta, and et~al.]{Barrett2024identifying}
Clark Barrett, Brad Boyd, Elie Burzstein, Nicholas Carlini, Brad Chen, Jihye Choi, Amrita~Roy Chowdhury, Mihai Christodorescu, Anupam Datta, and Soheil~Feizi et al.
\newblock \emph{{Identifying and Mitigating the Security Risks of Generative AI}}.
\newblock Now Foundations and Trends, 2024.

\bibitem[Bontcheva et~al.(2024)Bontcheva, Papadopoulous, Tsalakanidou, Gallotti, Dutkiewicz, Krack, Teyssou, Severio~Nucci, Spangenberg, Srba, et~al.]{bontcheva2024generative}
Kalina Bontcheva, Symeon Papadopoulous, Filareti Tsalakanidou, Riccardo Gallotti, Lidia Dutkiewicz, No{\'e}mie Krack, Denis Teyssou, Francesco Severio~Nucci, Jochen Spangenberg, Ivan Srba, et~al.
\newblock {Generative AI and disinformation: recent advances, challenges, and opportunities}.
\newblock 2024.

\bibitem[Chen et~al.(2024)Chen, Zeng, Yang, and Yang]{Chen2024drct}
Baoying Chen, Jishen Zeng, Jianquan Yang, and Rui Yang.
\newblock {DRCT: Diffusion Reconstruction Contrastive Training towards Universal Detection of Diffusion Generated Images}.
\newblock In \emph{ICML}, 2024.

\bibitem[Cheng et~al.(2025)Cheng, Lyu, Wang, Zhang, and Sehwag]{Cheng2025cospy}
Siyuan Cheng, Lingjuan Lyu, Zhenting Wang, Xiangyu Zhang, and Vikash Sehwag.
\newblock {CO-SPY: Combining Semantic and Pixel Features to Detect Synthetic Images by AI}.
\newblock In \emph{CVPR}, pages 13455--13465, 2025.

\bibitem[Corvi et~al.(2023)Corvi, Cozzolino, Zingarini, Poggi, Nagano, and Verdoliva]{Corvi2023detection}
Riccardo Corvi, Davide Cozzolino, Giada Zingarini, Giovanni Poggi, Koki Nagano, and Luisa Verdoliva.
\newblock {On the detection of synthetic images generated by diffusion models}.
\newblock In \emph{ICASSP}, pages 1--5, 2023.

\bibitem[Dang-Nguyen et~al.(2015)Dang-Nguyen, Pasquini, Conotter, and Boato]{nguyen2015raise}
Duc-Tien Dang-Nguyen, Cecilia Pasquini, Valentina Conotter, and Giulia Boato.
\newblock {RAISE: a raw images dataset for digital image forensics}.
\newblock In \emph{ACM Multimedia Systems Conference}, page 219–224. Association for Computing Machinery, 2015.

\bibitem[Dell’Anna et~al.(2025)Dell’Anna, Montibeller, and Boato]{DellAnna2025truefake}
Stefano Dell’Anna, Andrea Montibeller, and Giulia Boato.
\newblock {TrueFake: A Real World Case Dataset of Last Generation Fake Images also Shared on Social Networks}.
\newblock In \emph{International Joint Conference on Neural Networks (IJCNN)}, pages 1--8, 2025.

\bibitem[Dufour et~al.(2024)Dufour, Pathak, Samangouei, Hariri, Deshetti, Dudfield, Guess, Escayola, Tran, Babakar, et~al.]{Dufour2024ammeba}
Nicholas Dufour, Arkanath Pathak, Pouya Samangouei, Nikki Hariri, Shashi Deshetti, Andrew Dudfield, Christopher Guess, Pablo~Hern{\'a}ndez Escayola, Bobby Tran, Mevan Babakar, et~al.
\newblock {AMMeBa: A Large-Scale Survey and Dataset of Media-Based Misinformation In-The-Wild}.
\newblock \emph{arXiv preprint arXiv:2405.11697}, 2024.

\bibitem[Golestaneh et~al.(2022)Golestaneh, Dadsetan, and Kitani]{Golestaneh2022noreference}
S.~Alireza Golestaneh, Saba Dadsetan, and Kris~M. Kitani.
\newblock {No-Reference Image Quality Assessment via Transformers, Relative Ranking, and Self-Consistency}.
\newblock In \emph{WACV}, pages 1220--1230, 2022.

\bibitem[Guillaro et~al.(2025)Guillaro, Zingarini, Usman, Sud, Cozzolino, and Verdoliva]{Guillaro2025bfree}
Fabrizio Guillaro, Giada Zingarini, Ben Usman, Avneesh Sud, Davide Cozzolino, and Luisa Verdoliva.
\newblock {A Bias-Free Training Paradigm for More General AI-generated Image Detection}.
\newblock In \emph{CVPR}, pages 18685--18694, 2025.

\bibitem[Huang et~al.(2026)Huang, Lin, Tan, Du, Qiu, Zheng, Kong, Jiang, and Zheng]{Xia2025Mirage}
OuCheng Huang, Manxi Lin, Jiexiang Tan, Xiaoxiong Du, Yang Qiu, Junjun Zheng, Xiangheng Kong, Yuning Jiang, and Bo Zheng.
\newblock {MIRAGE: Towards AI-Generated Image Detection in the Wild}.
\newblock \emph{AAAI}, 40\penalty0 (7):\penalty0 5076--5084, 2026.

\bibitem[Huang et~al.(2025)Huang, Hu, Li, He, Zhao, Peng, Wu, Huang, and Cheng]{Huang2025sida}
Zhenglin Huang, Jinwei Hu, Xiangtai Li, Yiwei He, Xingyu Zhao, Bei Peng, Baoyuan Wu, Xiaowei Huang, and Guangliang Cheng.
\newblock {SIDA: Social Media Image Deepfake Detection, Localization and Explanation with Large Multimodal Model}.
\newblock In \emph{CVPR}, 2025.

\bibitem[Jiang et~al.(2025)Jiang, Yang, Chen, and Young]{Jiang2025new}
Jiajun Jiang, Wen-Chao Yang, Chung-Hao Chen, and Timothy Young.
\newblock {A New Deepfake Detection Method with No-Reference Image Quality Assessment to Resist Image Degradation}.
\newblock \emph{Eng}, 6\penalty0 (10), 2025.

\bibitem[Karageogiou et~al.(2024)Karageogiou, Bammey, Porcellini, Goupil, Teyssou, and Papadopoulos1]{Karageogiou2024evolution}
Dimitrios Karageogiou, Quentin Bammey, Valentin Porcellini, Bertrand Goupil, Denis Teyssou, and Symeon Papadopoulos1.
\newblock {Evolution of Detection Performance throughout the Online Lifespan of Synthetic Images}.
\newblock In \emph{Eur. Conf. Comput. Vis. Worksh.}, 2024.

\bibitem[Kim et~al.(2026)Kim, Lee, Park, and Kwon]{kim2024correlation}
Hyunjoon Kim, Jaehee Lee, Leo~Hyun Park, and Taekyoung Kwon.
\newblock On the correlation between deepfake detection performance and image quality metrics.
\newblock In \emph{3rd ACM Workshop on the Security Implications of Deepfakes and Cheapfakes (WDC)}, 2026.

\bibitem[Li et~al.(2025)Li, Wang, Li, Miao, Sun, Zhang, Ji, and Zhu]{Li2025bridging}
Chunxiao Li, Xiaoxiao Wang, Meiling Li, Boming Miao, Peng Sun, Yunjian Zhang, Xiangyang Ji, and Yao Zhu.
\newblock {Bridging the Gap Between Ideal and Real-world Evaluation: Benchmarking AI-Generated Image Detection in Challenging Scenarios}.
\newblock In \emph{ICCV}, 2025.

\bibitem[Lin et~al.(2024)Lin, Gupta, Zhang, Ren, Liu, Ding, Wang, Li, Verdoliva, and Hu]{Lin2024detecting}
Li Lin, Neeraj Gupta, Yue Zhang, Hainan Ren, Chun-Hao Liu, Feng Ding, Xin Wang, Xin Li, Luisa Verdoliva, and Shu Hu.
\newblock {Detecting multimedia generated by large AI models: A survey}.
\newblock \emph{arXiv preprint arXiv:2204.06125}, 2024.

\bibitem[Porcile et~al.(2023)Porcile, Gindi, Mundra, Verbus1, and Farid]{porcile2023finding}
Gonzalo J.~Aniano Porcile, Jack Gindi, Shivansh Mundra, James~R. Verbus1, and Hany Farid.
\newblock {Finding AI-Generated Faces in the Wild}.
\newblock In \emph{CVPR Workshops}, 2023.

\bibitem[Qui{\~{n}}onero-Candela et~al.(2006)Qui{\~{n}}onero-Candela, Rasmussen, Sinz, Bousquet, and Sch{\"o}lkopf]{Quinonero2006evaluating}
Joaquin Qui{\~{n}}onero-Candela, Carl~Edward Rasmussen, Fabian Sinz, Olivier Bousquet, and Bernhard Sch{\"o}lkopf.
\newblock {Evaluating Predictive Uncertainty Challenge}.
\newblock In \emph{Machine Learning Challenges. Evaluating Predictive Uncertainty, Visual Object Classification, and Recognising Textual Entailment}, pages 1--27. Springer Berlin Heidelberg, 2006.

\bibitem[Ricker et~al.(2024)Ricker, Assenmacher, Holz, Fischer, and Quiring]{ricker2024AI-generated}
Jonas Ricker, Dennis Assenmacher, Thorsten Holz, Asja Fischer, and Erwin Quiring.
\newblock {AI-Generated Faces in the Real World: A Large-Scale Case Study of Twitter Profile Images}.
\newblock In \emph{International Symposium on Research in Attacks, Intrusions and Defenses}, 2024.

\bibitem[Shin et~al.(2024)Shin, Lee, and Kim]{Shin2024blind}
Nyeong-Ho Shin, Seon-Ho Lee, and Chang-Su Kim.
\newblock {Blind Image Quality Assessment Based on Geometric Order Learning}.
\newblock In \emph{Proceedings of the IEEE/CVF Conference on Computer Vision and Pattern Recognition (CVPR)}, pages 12799--12808, 2024.

\bibitem[Song et~al.(2024)Song, Yan, Lin, Yao, Chen1, Chen, Zhao, Ding, and Li]{Song2024not}
Wentang Song, Zhiyuan Yan, Yuzhen Lin, Taiping Yao, Changsheng Chen1, Shen Chen, Yandan Zhao, Shouhong Ding, and Bin Li.
\newblock {Not All Fakes are Equal: A Quality-Centric Framework for Deepfake Detection}.
\newblock \emph{arXiv preprint arXiv:2411.05335v3}, 2024.

\bibitem[Tariang et~al.(2024)Tariang, Corvi, Cozzolino, Poggi, Nagano, and Verdoliva]{Tariang2024synthetic}
Diangarti Tariang, Riccardo Corvi, Davide Cozzolino, Giovanni Poggi, Koki Nagano, and Luisa Verdoliva.
\newblock {Synthetic Image Verification in the Era of Generative AI: What Works and What Isn't There Yet}.
\newblock \emph{IEEE Security \& Privacy}, 22:\penalty0 37--49, 2024.

\bibitem[Wang et~al.(2020)Wang, Wang, Zhang, Owens, and Efros]{Wang2020cnn}
Sheng-Yu Wang, Oliver Wang, Richard Zhang, Andrew Owens, and Alexei~A Efros.
\newblock {CNN-generated images are surprisingly easy to spot... for now}.
\newblock In \emph{CVPR}, pages 8695--8704, 2020.

\bibitem[Wu et~al.(2022)Wu, Zhou, Tian, and Liu]{Wu2022robust}
Haiwei Wu, Jiantao Zhou, Jinyu Tian, and Jun Liu.
\newblock {Robust image forgery detection over online social network shared images}.
\newblock In \emph{CVPR}, 2022.

\bibitem[Xu et~al.(2024)Xu, Liao, Xiao, Chen, Wu, Yan, and Lin]{Xu2024boosting}
Kangmin Xu, Liang Liao, Jing Xiao, Chaofeng Chen, Haoning Wu, Qiong Yan, and Weisi Lin.
\newblock {Boosting Image Quality Assessment through Efficient Transformer Adaptation with Local Feature Enhancement}.
\newblock In \emph{CVPR}, pages 2662--2672, 2024.

\bibitem[Yang et~al.(2025)Yang, Qian, Zhu, Russakovsky, and Wu]{Yang2025d3}
Yongqi Yang, Zhihao Qian, Ye Zhu, Olga Russakovsky, and Yu Wu.
\newblock {D{\textasciicircum}3: Scaling Up Deepfake Detection by Learning from Discrepancy}.
\newblock In \emph{CVPR}, pages 23850--23859, 2025.

\bibitem[Zhan et~al.(2021)Zhan, Yu, Wu, Zhang, Lu, Liu, Kortylewski, Theobalt, and Xing]{Zhan2023multimodal}
Fangneng Zhan, Yingchen Yu, Rongliang Wu, Jiahui Zhang, Shijian Lu, Lingjie Liu, Adam Kortylewski, Christian Theobalt, and Eric Xing.
\newblock {Multimodal Image Synthesis and Editing: The Generative AI Era}.
\newblock \emph{IEEE TPAMI}, 45\penalty0 (12):\penalty0 15098--15119, 2021.

\end{thebibliography}
}

\end{document}